\documentclass[conference]{IEEEtran}
\usepackage{cite}
\usepackage{amsmath,amssymb,amsfonts}
\usepackage{algorithmic}
\usepackage{graphicx}
\usepackage{textcomp}
\usepackage{xcolor}
\def\BibTeX{{\rm B\kern-.05em{\sc i\kern-.025em b}\kern-.08em
    T\kern-.1667em\lower.7ex\hbox{E}\kern-.125emX}}

\begin{document}

\title{Indoor Navigation Algorithm Based on a Smartphone Inertial Measurement Unit and Map Matching}

\author{\IEEEauthorblockN{Taewon Kang}
\IEEEauthorblockA{\textit{School of Integrated Technology} \\
\textit{Yonsei University}\\
Incheon, Republic of Korea \\
taewon.kang@yonsei.ac.kr}
\and
\IEEEauthorblockN{Younghoon Shin${}^{*}$}
\IEEEauthorblockA{\textit{School of Integrated Technology} \\
\textit{Yonsei University}\\
Incheon, Republic of Korea \\
yh.s@yonsei.ac.kr} 
{\small${}^{*}$ Corresponding author}
}

\maketitle

\begin{abstract}
We propose an indoor navigation algorithm based on pedestrian dead reckoning (PDR) using an inertial measurement unit in a smartphone and map matching. The proposed indoor navigation system is user-friendly and convenient because it requires no additional device except a smartphone and works with a pedestrian in a casual posture who is walking with a smartphone in their hand. Because the performance of the PDR decreases over time, we greatly reduced the position error of the trajectory estimated by PDR using a map matching method with a known indoor map. To verify the proposed indoor navigation algorithm, we conducted an experiment in a real indoor environment using a commercial Android smartphone. The performance of our algorithm was demonstrated through the results of the experiment.
\end{abstract}

\begin{IEEEkeywords}
indoor navigation, pedestrian dead reckoning (PDR), map matching, inertial measurement unit (IMU)
\end{IEEEkeywords}

\section{Introduction}
Because the indoor environment has been broadened and complexified recently, the demand for indoor location information has been increasing for indoor wireless communication applications, such as location-based services (LBS) \cite{Yassin20171327, Gresmann2010107}. Despite several navigation systems that use wireless signals such as global navigation satellite systems (GNSS) \cite{Enge11, Yoon2016, Kim201414971, Saito20171937, Yoon2020, Sun21:Markov, Sun2020889, Park2021919, Park2018387, Kim19:Mitigation}, enhanced long-range navigation (eLoran) \cite{Lo201023, Son2018666, Son20191828, Kim2020796, Park2020824, Son20181034, Rhee21:Enhanced}, and other systems \cite{Kim2017:SFOL, Seo20142224, Rhee2019, Rhee2018224, Shin2017617, Kang20191182, Jia21:Ground, Lee2020:Preliminary, Jeong2020958, Lee2020939, Lee20202347}, they are not suitable for indoor positioning because of signal blockage, reflection, or interference in indoor environments \cite{Kang2020774, Kang20191182}.

Pedestrian dead reckoning (PDR) \cite{Li201524862, Zhuang2015793, Xiao2014187} is one of the widely used methods for indoor navigation because of its low power requirements, cost-effectiveness, and availability without any infrastructure. In general, PDR is achieved with the aid of an inertial measurement unit (IMU), a combination of accelerometers and gyroscopes. The most accessible IMU sensor in our everyday life is contained in smartphones; hence, many PDR algorithms have been implemented using smartphones \cite{Wang2016, Chen201524595, Park2020800, Han19:Smartphone, Zhao2019}. However, because PDR is vulnerable to cumulative errors over time, it is known to be insufficient for use solely in indoor navigation systems. To reduce its errors, different approaches to aid PDR have been implemented, such as Wi-Fi-based positioning system (WPS) \cite{Jeon2014385}, magnetic matching (MM) method \cite{Li2016169}, and vision-aided method \cite{Yan20181}. However, these methods require additional measuring device or database. Another promising approach is the map matching method, which can aid PDR by periodically alleviating cumulative errors \cite{Perttula20142682, Zampella20151304, Aggarwal2011}. With known indoor environment information, such as a floor plan, the position and heading of an inaccurate trajectory can be corrected. Impossible pedestrian positions, such as a position inside a pillar, can be excluded from the possible position candidates. 
Khalifa and Hassan \cite{Khalifa2012} suggested activity-based map matching, which is comprised of detecting what a person is doing and matching the position on the map where the detected activity is possible. 
Kamiya \textit{et al.} \cite{Kamiya2019} proposed context-based map matching, by distinguishing the different movement context and matching the user position to the indoor road network.

In this study, we developed an indoor navigation algorithm based on PDR using an IMU in a smartphone and map matching. The proposed method can be implemented by a typical pedestrian in a casual posture who is walking with a smartphone in their hand. Using the data collected with a smartphone IMU, the rough trajectory of a pedestrian is estimated by PDR. The trajectory is corrected using map matching, which greatly reduces the position error of the PDR. The methodology of the proposed indoor navigation system is discussed in Section \ref{sec:Methodology}. The experimental environment and results are presented in Section \ref{sec:Experiment}. Finally, conclusions are presented in Section \ref{sec:Conclusion}.

\section{Methodology}
\label{sec:Methodology}

\subsection{Pedestrian dead reckoning}
PDR was used to estimate the approximate trajectory \cite{Moon2019157, Moon20181530} of a pedestrian in our method. The PDR consists of three parts: step detection, heading estimation, and position estimation. First, step detection was performed by detecting the peaks in the accelerometer signal data. The heading of the pedestrian is estimated using a gyroscope. Because the initial heading is not correctly synchronized with the actual indoor trajectory, the position error is biased and accumulates over time. The positions of each step are then estimated by adding double-integrated acceleration values from the previous position, in both the horizontal and vertical directions, as in:
\begin{equation}
\begin{pmatrix}
X_{k+1,P\!D\!R} \\
Y_{k+1,P\!D\!R}
\end{pmatrix}
=
\begin{pmatrix}
X_{k,P\!D\!R} \\
Y_{k,P\!D\!R}
\end{pmatrix}
+
\begin{pmatrix}
l \cos \varphi  \\
l  \sin \varphi
\end{pmatrix} \label{eq1}
\end{equation}
where $\begin{pmatrix}
X_{k+1,P\!D\!R} \\
Y_{k+1,P\!D\!R}
\end{pmatrix}$ and $\begin{pmatrix}
X_{k,P\!D\!R} \\
Y_{k,P\!D\!R}
\end{pmatrix}$ represent the position coordinates of $(k+1)$ and $k$-th step estimated by PDR, respectively, $l$ is the step length, and $\varphi$ is the heading angle. The initial position of the pedestrian is assumed to be known.

\subsection{Map matching}
Map matching was used to match the trajectory estimated by PDR in our proposed method. Because map matching is performed in accordance with the actual map or coordinates, map matching can rearrange the trajectory and prevent the estimated position from being located in incorrect or impossible places. We used the points at which the turn occurred in the trajectory for the matching points. The turn points in the PDR trajectory can be detected by detecting the step in which the heading angle exceeds a predefined threshold value. In an indoor environment, turns occur at the corner of the corridor. Hence, it is reasonable to match the turn points in the trajectory to the corner points of the indoor map. To perform map matching, we divided the trajectory with respect to the estimated turn points, and we matched the turn points to the corner point coordinates. The step points between the turn points were shifted and rotated with respect to the previous and next turn points, as in:

\begin{equation}
\begin{split}
\begin{pmatrix}
X_{k,mm} \\
Y_{k,mm}
\end{pmatrix}
& =
\begin{pmatrix}
X_{k,P\!D\!R}-X_{tp} \\
Y_{k,P\!D\!R}-Y_{tp}
\end{pmatrix}
\times \cos \theta \\
& +
\begin{pmatrix}
Y_{k,P\!D\!R}-Y_{tp} \\
-X_{k,P\!D\!R}+X_{tp}
\end{pmatrix}
\times \sin \theta \\
& +
\begin{pmatrix}
X_{tp} \\
Y_{tp}
\end{pmatrix} \label{eq2}
\end{split}
\end{equation}
where
$\begin{pmatrix}
X_{k,mm} \\
Y_{k,mm}
\end{pmatrix}$ represent the position coordinates of $k$th step estimated by map matching, $\begin{pmatrix}
X_{tp} \\
Y_{tp}
\end{pmatrix}$ represent the turn point coordinates, and $\theta$ represent the angle between the line connecting the previous and next turn points and the line connecting the matching points estimated by PDR.

\section{Experimental Environment and Results}
\label{sec:Experiment}

\subsection{Experimental environment}
\begin{figure}
\centerline{\includegraphics[width=1\linewidth]{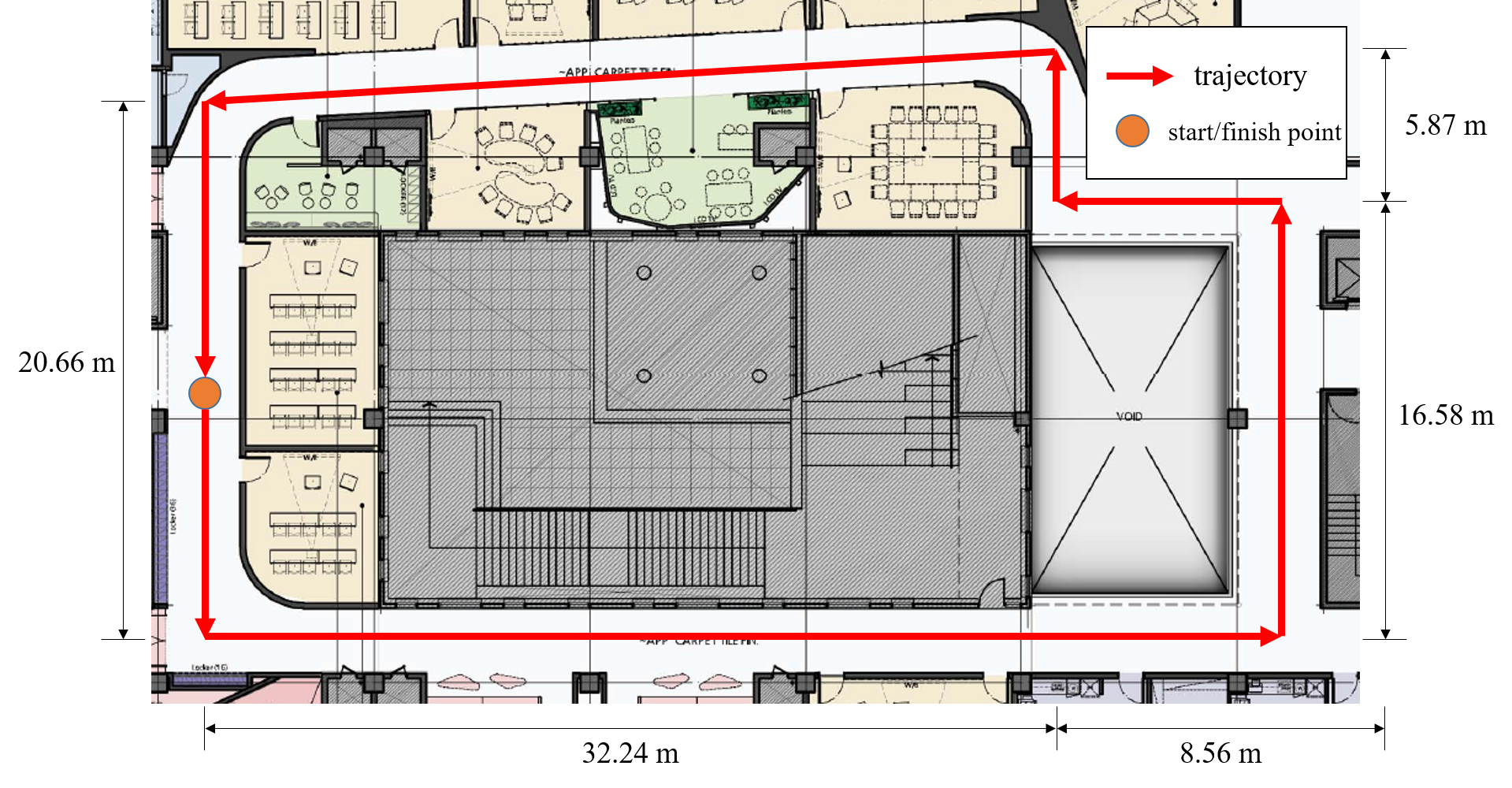}}
\caption{Floor plan of the test site and the experimental trajectory.}
\label{fig:FloorPlan}
\end{figure}

To verify the proposed system, we conducted an IMU data collection experiment on the second floor of the Veritas Hall C building (Yonsei University, Incheon, South Korea). Fig.~\ref{fig:FloorPlan} shows the floor plan of the test site and the trajectory of the data collection experiment. To collect the data, we used the Sensor Tester version 2.2.2 application installed on a Samsung Galaxy S8 smartphone. The experimenter walked through the experimental trajectory with a casual posture, carrying a smartphone and watching its screen. The experimental trajectory was approximately 125 m.

\subsection{Experimental results}

\begin{figure}
\centerline{\includegraphics[width=1\linewidth]{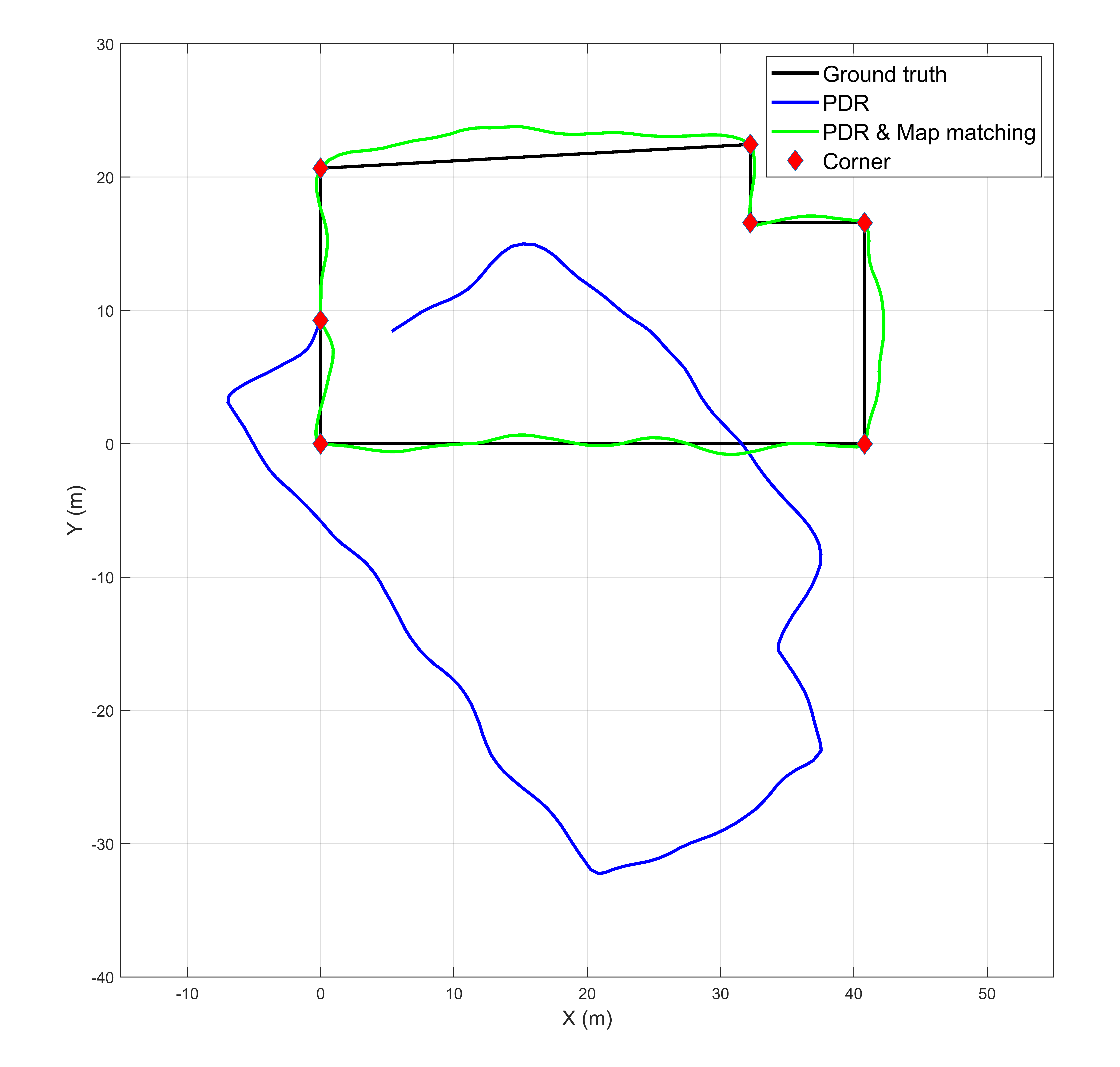}}
\caption{Actual trajectory and the trajectories of PDR and the proposed method.}
\label{fig:Trajectory}
\end{figure}

Fig.~\ref{fig:Trajectory} shows the trajectory estimated by PDR only and that determined by the proposed method from the experiment. The results show that the trajectory estimated by PDR only has a large position error compared with the ground truth path because the initial heading is not correctly estimated. It should be noted that it is difficult to estimate the initial heading using inertial sensors. Although the start and finish points of the experimental path are the same, the start and finish points of the PDR-only trajectory are 5.4 m away, demonstrating that the positioning error accumulates over time. From the PDR-only trajectory, the points at which the turn occurs are detected, and these are matched with the actual corner points of the experimental path, which are indicated in red diamonds in Fig.~\ref{fig:Trajectory}. The green curve in Fig.~\ref{fig:Trajectory} after the map matching process follows the ground truth trajectory.

\begin{figure}
\centerline{\includegraphics[width=1\linewidth]{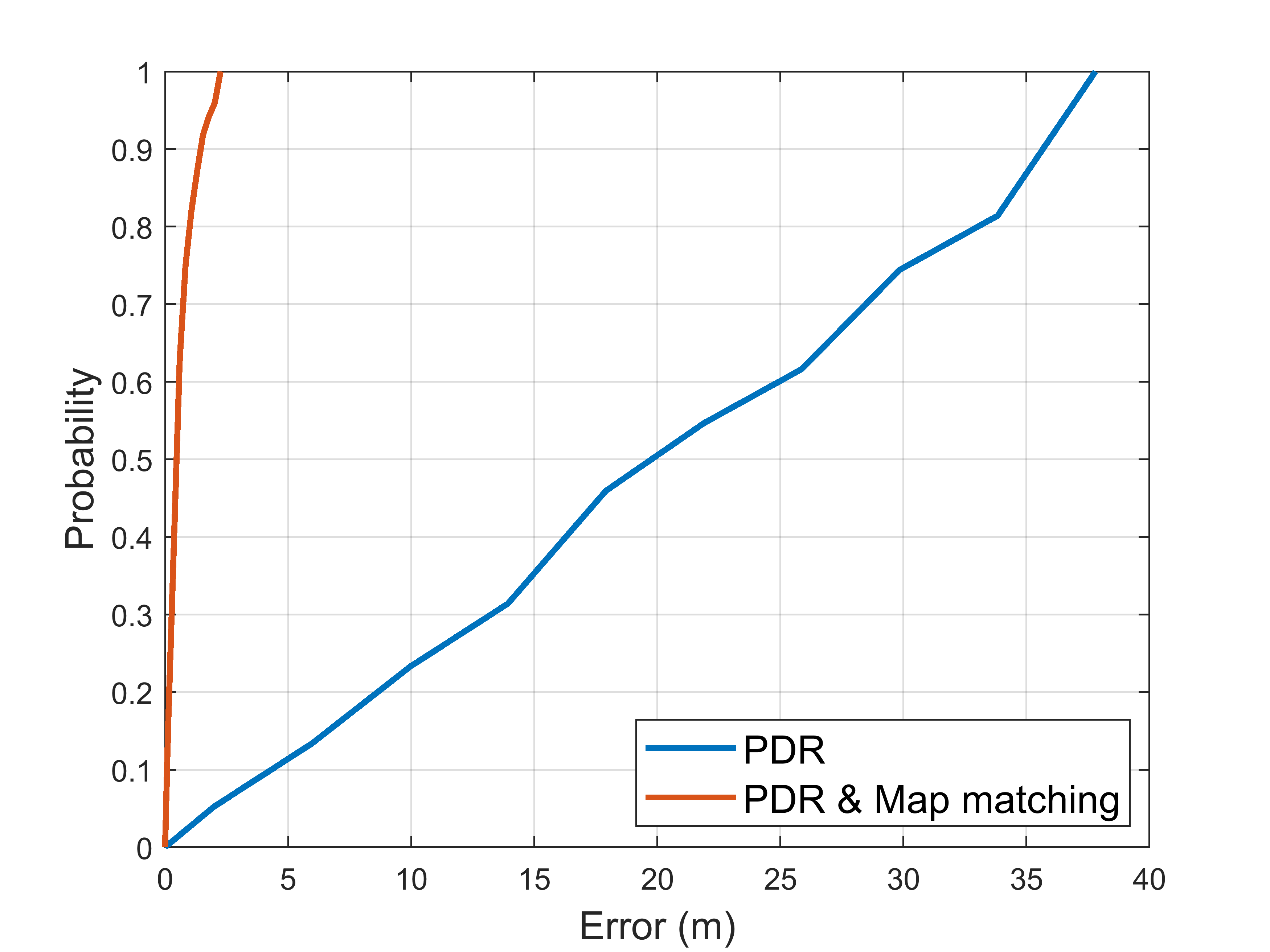}}
\caption{Position error CDF.}
\label{fig:CDF}
\end{figure}

The position error cumulative distribution function (CDF) is shown in Fig.~\ref{fig:CDF}. The maximum position error of the proposed method in the experiment is 2.4 m, which is a far smaller value than the maximum position error of the PDR-only method (39.8 m).

Table~\ref{tab:Performance} lists the statistical results of the experiment by comparing PDR-only and the proposed method. The proposed method reduces the position errors by 96.9\% compared with the PDR-only method. 

\begin{table}
\caption{Positioning performance of PDR-only and the proposed method in the experiment}
\begin{center}
\begin{tabular}{|c|c|c|}
\hline
\textbf{Method}     & \textbf{Mean error (m)} & \textbf{Standard deviation (m)} \\ \hline
PDR-only  & 22.4    & 11.7                           \\ \hline
PDR and map matching & 0.7    & 0.6                            \\ \hline
\end{tabular}
\label{tab:Performance}
\end{center}
\end{table}

\section{Conclusion}
\label{sec:Conclusion}

An indoor navigation algorithm based on PDR using an IMU in a smartphone and map matching was presented. The proposed system does not require additional infrastructure and works with a pedestrian in a casual posture who is walking with a smartphone in his or her hand. PDR is used first to estimate the trajectory of a pedestrian, and map matching is conducted to correct the rough trajectory with known indoor environment information. From the experiment conducted in a real indoor environment, the proposed method was demonstrated to reduce the position error of the trajectory estimated by PDR solely by 96.9\%. The experimental results show that map matching can successfully aid PDR, and the proposed method provides accurate positioning.

\section*{Acknowledgment}

This work was supported by Institute for Information \& Communications Technology Planning \& Evaluation (IITP) grant funded by the Korea government (KNPA) (No. 2019-0-01291, LTE-based accurate positioning technique for emergency rescue).

\bibliographystyle{IEEEtran}
\bibliography{mybibfile, IUS_publications}

\begin{thebibliography}{10}
\providecommand{\url}[1]{#1}
\csname url@samestyle\endcsname
\providecommand{\newblock}{\relax}
\providecommand{\bibinfo}[2]{#2}
\providecommand{\BIBentrySTDinterwordspacing}{\spaceskip=0pt\relax}
\providecommand{\BIBentryALTinterwordstretchfactor}{4}
\providecommand{\BIBentryALTinterwordspacing}{\spaceskip=\fontdimen2\font plus
\BIBentryALTinterwordstretchfactor\fontdimen3\font minus
  \fontdimen4\font\relax}
\providecommand{\BIBforeignlanguage}[2]{{%
\expandafter\ifx\csname l@#1\endcsname\relax
\typeout{** WARNING: IEEEtran.bst: No hyphenation pattern has been}%
\typeout{** loaded for the language `#1'. Using the pattern for}%
\typeout{** the default language instead.}%
\else
\language=\csname l@#1\endcsname
\fi
#2}}
\providecommand{\BIBdecl}{\relax}
\BIBdecl

\bibitem{Yassin20171327}
A.~Yassin, Y.~Nasser, M.~Awad, A.~Al-Dubai, R.~Liu, C.~Yuen, R.~Raulefs, and
  E.~Aboutanios, ``Recent advances in indoor localization: A survey on
  theoretical approaches and applications,'' \emph{IEEE Commun. Surv. Tutor.},
  vol.~19, no.~2, pp. 1327--1346, 2017.

\bibitem{Gresmann2010107}
B.~Greßmann, H.~Klimek, and V.~Turau, ``Towards ubiquitous indoor location
  based services and indoor navigation,'' in \emph{Proc. WPNC}, 2010, pp.
  107--112.

\bibitem{Enge11}
P.~Misra and P.~Enge, \emph{Global Positioning System: Signals, Measurements,
  and Performance}.\hskip 1em plus 0.5em minus 0.4em\relax Ganga-Jamuna Press,
  2011.

\bibitem{Yoon2016}
D.~Yoon, C.~Kee, J.~Seo, and B.~Park, ``Position accuracy improvement by
  implementing the {DGNSS-CP} algorithm in smartphones,'' \emph{Sensors},
  vol.~16, no.~6, Jun. 2016.

\bibitem{Kim201414971}
M.~Kim, J.~Seo, and J.~Lee, ``A comprehensive method for {GNSS} data quality
  determination to improve ionospheric data analysis,'' \emph{Sensors},
  vol.~14, no.~8, pp. 14\,971--14\,993, Aug. 2014.

\bibitem{Saito20171937}
S.~Saito, S.~Sunda, J.~Lee, S.~Pullen, S.~Supriadi, T.~Yoshihara,
  M.~Terkildsen, F.~Lecat, and {ICAO APANPIRG Ionospheric Studies Task Force},
  ``Ionospheric delay gradient model for {GBAS} in the {Asia-Pacific} region,''
  \emph{GPS Solut.}, vol.~21, no.~4, pp. 1937--1947, 2017.

\bibitem{Yoon2020}
H.~Yoon, H.~Seok, C.~Lim, and B.~Park, ``An online {SBAS} service to improve
  drone navigation performance in high-elevation masked areas,''
  \emph{Sensors}, vol.~20, no.~11, 2020.

\bibitem{Sun21:Markov}
A.~K. Sun, H.~Chang, S.~Pullen, H.~Kil, J.~Seo, Y.~J. Morton, and J.~Lee,
  ``Markov chain-based stochastic modeling of deep signal fading: Availability
  assessment of dual-frequency {GNSS}-based aviation under ionospheric
  scintillation,'' Space Weather, in press.

\bibitem{Sun2020889}
K.~Sun, H.~Chang, J.~Lee, J.~Seo, Y.~Jade~Morton, and S.~Pullen, ``Performance
  benefit from dual-frequency {GNSS}-based aviation applications under
  ionospheric scintillation: {A} new approach to fading process modeling,'' in
  \emph{Proc. ION ITM}, Jan. 2020, pp. 889--899.

\bibitem{Park2021919}
K.~Park and J.~Seo, ``Single-antenna-based {GPS} antijamming method exploiting
  polarization diversity,'' \emph{IEEE Trans. Aerosp. Electron. Syst.},
  vol.~57, no.~2, pp. 919--934, Apr. 2021.

\bibitem{Park2018387}
K.~Park, D.~Lee, and J.~Seo, ``Dual-polarized {GPS} antenna array algorithm to
  adaptively mitigate a large number of interference signals,'' \emph{Aerosp.
  Sci. Technol.}, vol.~78, pp. 387--396, Jul. 2018.

\bibitem{Kim19:Mitigation}
S.~Kim, K.~Park, and J.~Seo, ``Mitigation of {GPS} chirp jammer using a
  transversal {FIR} filter and {LMS} algorithm,'' in \emph{Proc. ITC-CSCC},
  Jun. 2019.

\bibitem{Lo201023}
S.~Lo, B.~Peterson, T.~Hardy, and P.~Enge, ``Improving {Loran} coverage with
  low power transmitters,'' \emph{J. Navig.}, vol.~63, no.~1, pp. 23--38, 2010.

\bibitem{Son2018666}
P.-W. Son, J.~Rhee, and J.~Seo, ``Novel multichain-based {Loran} positioning
  algorithm for resilient navigation,'' \emph{IEEE Trans. Aerosp. Electron.
  Syst.}, vol.~54, no.~2, pp. 666--679, Oct. 2018.

\bibitem{Son20191828}
P.-W. Son, J.~Rhee, J.~Hwang, and J.~Seo, ``Universal kriging for {Loran} {ASF}
  map generation,'' \emph{IEEE Trans. Aerosp. Electron. Syst.}, vol.~55, no.~4,
  pp. 1828--1842, Oct. 2019.

\bibitem{Kim2020796}
W.~Kim, P.-W. Son, J.~Rhee, and J.~Seo, ``Development of record and management
  software for {GPS}/{Loran} measurements,'' in \emph{Proc. ICCAS}, Oct. 2020,
  pp. 796--799.

\bibitem{Park2020824}
J.~Park, P.-W. Son, W.~Kim, J.~Rhee, and J.~Seo, ``Effect of outlier removal
  from temporal {ASF} corrections on multichain {Loran} positioning accuracy,''
  in \emph{Proc. ICCAS}, Oct. 2020, pp. 824--826.

\bibitem{Son20181034}
P.-W. Son, J.~Rhee, Y.~Han, K.~Seo, and J.~Seo, ``Preliminary study of
  multichain-based {Loran} positioning accuracy for a dynamic user in {South
  Korea},'' in \emph{Proc. IEEE/ION PLANS}, Apr. 2018, pp. 1034--1038.

\bibitem{Rhee21:Enhanced}
J.~H. Rhee, S.~Kim, P.-W. Son, and J.~Seo, ``Enhanced accuracy simulator for a
  future {Korean} nationwide {eLoran} system,'' IEEE Access, in press.

\bibitem{Kim2017:SFOL}
E.~Kim and J.~Seo, ``{SFOL} pulse: {A} high accuracy {DME} pulse for
  alternative aircraft position and navigation,'' \emph{Sensors}, vol.~17,
  no.~10, Sep. 2017.

\bibitem{Seo20142224}
J.~Seo and T.~Walter, ``Future dual-frequency {GPS} navigation system for
  intelligent air transportation under strong ionospheric scintillation,''
  \emph{IEEE Trans. Intell. Transp. Syst.}, vol.~15, no.~5, pp. 2224--2236,
  Apr. 2014.

\bibitem{Rhee2019}
J.~Rhee and J.~Seo, ``Low-cost curb detection and localization system using
  multiple ultrasonic sensors,'' \emph{Sensors}, vol.~19, no.~6, Mar. 2019.

\bibitem{Rhee2018224}
------, ``Ground reflection elimination algorithms for enhanced distance
  measurement to the curbs using ultrasonic sensors,'' in \emph{Proc. ION ITM},
  Jan. 2018, pp. 224--231.

\bibitem{Shin2017617}
Y.~Shin, S.~Lee, and J.~Seo, ``Autonomous safe landing-area determination for
  rotorcraft {UAVs} using multiple {IR-UWB} radars,'' \emph{Aerosp. Sci.
  Technol.}, vol.~69, pp. 617--624, Oct. 2017.

\bibitem{Kang20191182}
T.~Kang, H.~Lee, and J.~Seo, ``Analysis of the maximum correlation peak value
  and {RSRQ} in {LTE} signals according to frequency bands and sampling
  frequencies,'' in \emph{Proc. ICCAS}, Oct. 2019, pp. 1182--1186.

\bibitem{Jia21:Ground}
M.~Jia, H.~Lee, J.~Khalife, Z.~M. Kassas, and J.~Seo, ``Ground vehicle
  navigation integrity monitoring for multi-constellation {GNSS} fused with
  cellular signals of opportunity,'' in \emph{Proc. IEEE ITSC}, 2021.

\bibitem{Lee2020:Preliminary}
H.~Lee and J.~Seo, ``A preliminary study of machine-learning-based ranging with
  {LTE} channel impulse response in multipath environment,'' in \emph{Proc.
  IEEE ICCE-Asia}, Nov. 2020.

\bibitem{Jeong2020958}
S.~Jeong, H.~Lee, T.~Kang, and J.~Seo, ``{RSS}-based {LTE} base station
  localization using single receiver in environment with unknown path-loss
  exponent,'' in \emph{Proc. ICTC}, Oct. 2020, pp. 958--961.

\bibitem{Lee2020939}
H.~Lee, A.~Abdallah, J.~Park, J.~Seo, and Z.~Kassas, ``Neural network-based
  ranging with {LTE} channel impulse response for localization in indoor
  environments,'' in \emph{Proc. ICCAS}, Oct. 2020, pp. 939--944.

\bibitem{Lee20202347}
H.~Lee, J.~Seo, and Z.~Kassas, ``Integrity-based path planning strategy for
  urban autonomous vehicular navigation using {GPS} and cellular signals,'' in
  \emph{Proc. ION GNSS+}, Sep. 2020, pp. 2347--2357.

\bibitem{Kang2020774}
T.~Kang and J.~Seo, ``Practical simplified indoor multiwall path-loss model,''
  in \emph{Proc. ICCAS}, Oct. 2020, pp. 774--777.

\bibitem{Li201524862}
X.~Li, J.~Wang, and C.~Liu, ``A bluetooth/{PDR} integration algorithm for an
  indoor positioning system,'' \emph{Sensors}, vol.~15, no.~10, pp.
  24\,862--24\,885, 2015.

\bibitem{Zhuang2015793}
Y.~Zhuang, H.~Lan, Y.~Li, and N.~El-Sheimy, ``{PDR/INS/WiFi} integration based
  on handheld devices for indoor pedestrian navigation,'' \emph{Micromachines},
  vol.~6, no.~6, pp. 793--812, 2015.

\bibitem{Xiao2014187}
Z.~Xiao, H.~Wen, A.~Markham, and N.~Trigoni, ``Robust pedestrian dead reckoning
  ({R-PDR}) for arbitrary mobile device placement,'' in \emph{Proc. IPIN},
  2014, pp. 187--196.

\bibitem{Wang2016}
X.~Wang, M.~Jiang, Z.~Guo, N.~Hu, Z.~Sun, and J.~Liu, ``An indoor positioning
  method for smartphones using landmarks and {PDR},'' \emph{Sensors}, vol.~16,
  no.~12, 2016.

\bibitem{Chen201524595}
G.~Chen, X.~Meng, Y.~Wang, Y.~Zhang, P.~Tian, and H.~Yang, ``Integrated
  {WiFi/PDR}/smartphone using an unscented {Kalman} filter algorithm for {3D}
  indoor localization,'' \emph{Sensors}, vol.~15, no.~9, pp. 24\,595--24\,614,
  2015.

\bibitem{Park2020800}
K.~Park, W.~Kim, and J.~Seo, ``Effects of initial attitude estimation errors on
  loosely coupled smartphone {GPS/IMU} integration system,'' in \emph{Proc.
  ICCAS}, Oct. 2020, pp. 800--803.

\bibitem{Han19:Smartphone}
S.~Han, T.~Kang, and J.~Seo, ``Smartphone application to estimate distances
  from {LTE} base stations based on received signal strength measurements,'' in
  \emph{Proc. ITC-CSCC}, Jun. 2019.

\bibitem{Zhao2019}
H.~Zhao, W.~Cheng, N.~Yang, S.~Qiu, Z.~Wang, and J.~Wang, ``Smartphone-based
  {3D} indoor pedestrian positioning through multi-modal data fusion,''
  \emph{Sensors}, vol.~19, no.~20, 2019.

\bibitem{Jeon2014385}
S.~Jeon, J.~Lee, H.~Hong, S.~Shin, and H.~Lee, ``Indoor {WPS/PDR} performance
  enhancement using map matching algorithm with mobile phone,'' in \emph{Proc.
  IEEE PLANS}, 2014, pp. 385--392.

\bibitem{Li2016169}
Y.~Li, Y.~Zhuang, H.~Lan, Q.~Zhou, X.~Niu, and N.~El-Sheimy, ``A hybrid
  {WiFi}/magnetic matching/{PDR} approach for indoor navigation with smartphone
  sensors,'' \emph{IEEE Commun. Lett.}, vol.~20, no.~1, pp. 169--172, 2016.

\bibitem{Yan20181}
J.~Yan, G.~He, A.~Basiri, and C.~Hancock, ``Vision-aided indoor pedestrian dead
  reckoning,'' in \emph{Proc. IEEE I2MTC}, 2018, pp. 1--6.

\bibitem{Perttula20142682}
A.~Perttula, H.~Leppakoski, M.~Kirkko-Jaakkola, P.~Davidson, J.~Collin, and
  J.~Takala, ``Distributed indoor positioning system with inertial measurements
  and map matching,'' \emph{IEEE Trans. Instrum. Meas.}, vol.~63, no.~11, pp.
  2682--2695, 2014.

\bibitem{Zampella20151304}
F.~Zampella, A.~Jimenez~Ruiz, and F.~Seco~Granja, ``Indoor positioning using
  efficient map matching, {RSS} measurements, and an improved motion model,''
  \emph{IEEE Trans. Veh. Technol.}, vol.~64, no.~4, pp. 1304--1317, 2015.

\bibitem{Aggarwal2011}
P.~Aggarwal, D.~Thomas, L.~Ojeda, and J.~Borenstein, ``Map matching and
  heuristic elimination of gyro drift for personal navigation systems in
  {GPS}-denied conditions,'' \emph{Meas. Sci. Technol.}, vol.~22, no.~2, 2011.

\bibitem{Khalifa2012}
S.~Khalifa and M.~Hassan, ``Evaluating mismatch probability of activity-based
  map matching in indoor positioning,'' in \emph{Proc. IPIN}, 2012, pp. 1--9.

\bibitem{Kamiya2019}
Y.~Kamiya, Y.~Gu, and S.~Kamijo, ``Indoor positioning in large shopping mall
  with context based map matching,'' in \emph{Proc. IEEE ICCE}, 2019, pp. 1--6.

\bibitem{Moon2019157}
H.-S. Moon and J.~Seo, ``Prediction of human trajectory following a haptic
  robotic guide using recurrent neural networks,'' in \emph{Proc. IEEE WHC},
  Aug. 2019, pp. 157--162.

\bibitem{Moon20181530}
H.-S. Moon, W.~Kim, S.~Han, and J.~Seo, ``Observation of human trajectory in
  response to haptic feedback from mobile robot,'' in \emph{Proc. ICCAS}, Oct.
  2018, pp. 1530--1534.

\end{thebibliography}

\vspace{12pt}

\end{document}